\pdfoutput=1

\documentclass[11pt]{article}

\usepackage{naacl2021}

\usepackage{times}
\usepackage{latexsym}
\usepackage{graphicx}
\usepackage{multirow}
\usepackage{latexsym}
\usepackage{bbm}

\usepackage{CJK}
\usepackage{verbatim}
\usepackage{url}
\usepackage{footnote}
\usepackage{bbm}
\usepackage{amsfonts}
\usepackage{amsmath}
\usepackage{algpseudocode}
\usepackage{amsmath}
\usepackage{url}
\usepackage[T1]{fontenc}

\usepackage[utf8]{inputenc}

\usepackage{microtype}

\title{BSTC: A Large-Scale Chinese-English Speech Translation Dataset}

\author{Ruiqing Zhang, Xiyang Wang, Chuanqiang Zhang, Zhongjun He \\ \textbf{Hua Wu, Zhi Li, Haifeng Wang, Ying Chen, Qinfei Li}\\
  Baidu Inc. No. 10, Shangdi 10th Street, Beijing, 100085, China \\
  {\{zhangruiqing01, zhangchuanqiang, hezhongjun, wu\_hua\}@baidu.com} 
}

\date{}

\begin{document}
\begin{CJK*}{UTF8}{gbsn}
\maketitle

\begin{abstract}
This paper presents BSTC (Baidu Speech Translation Corpus), a large-scale Chinese-English speech translation dataset. This dataset is constructed based on a collection of licensed videos of talks or lectures, including about 68 hours of Mandarin data, their manual transcripts and translations into English, as well as automated transcripts by an automatic speech recognition (ASR) model. We have further asked three experienced interpreters to simultaneously interpret the testing talks in a mock conference setting. This corpus is expected to promote the research of automatic simultaneous translation as well as the development of practical systems. We have organized simultaneous translation tasks and used this corpus to evaluate automatic simultaneous translation systems.

\end{abstract}
\section{Introduction}
In recent years, automatic speech translation (AST) has attracted increasing interest for its commercial potential ({\em e.g.}, {\em Simultaneous Interpretation} and {\em Wireless Speech Translator}). A large amount of research has focused on speech translation \cite{weiss2017sequence,niehues2018,chung2018unsupervised,sperber19tacl,kahn2020libri,inaguma2020espnet} and simultaneous translation \cite{sridhar2013segmentation,oda2014optimizing,cho2016can,gu2017learning,DBLP:journals/corr/abs-1810-08398,arivazhagan2019monotonic,zhang-etal-2020-learning-adaptive}. The former intends to convert speech signals in the source language to the target language, and the latter aims to achieve a real-time translation that delivers the speech to the audience in the target language while minimizing the delay between the speaker and the translation.

\begin{table}[t]
\begin{center}
\resizebox{\linewidth}{!}{
\begin{tabular}{l|c|c}
\hline\hline
\textit{\textbf{Speech Translation}} & \textbf{Languages} & \textbf{Hours} \\
\hline
F-C \shortcite{post2013improved} & Es$\rightarrow$En & 38\\
KIT-Disfluency \shortcite{cho-etal-2014-corpus} &De$\rightarrow$En & 13 \\
BTEC \shortcite{berard2016listen} & En$\rightarrow$Fr & 17 \\
MSLT V1.0 \shortcite{federmann2016microsoft} & En$\leftrightarrow$Fr/De & 23 \\
\multirow{3}{*}{MSLT V1.1 \shortcite{federmann2017microsoft}} & En$\rightarrow$Zh/Jp & 6 \\
& Zh$\rightarrow$En & 5 \\
& Jp $\rightarrow$En & 9 \\
Travel \shortcite{woldeyohannis2017corpus} &Am$\rightarrow$En & 8 \\
Aug-LibriSpeech \shortcite{kocabiyikoglu2018augmenting} &En$\rightarrow$Fr &236 \\
MuST-C \shortcite{di2019must} & En$\rightarrow$8 Euro langs & 3617 \\
Europarl-ST \shortcite{iranzo2020europarl} & 9 Euro langs & 1642\\
Covost \shortcite{wang2020covost,wang2020covost2} & En$\leftrightarrow$21 langs & 2880 \\
\hline\hline
\textit{\textbf{Simultaneous Translation}} & \textbf{Languages} & \textbf{Hours} \\\hline
CIAIR \shortcite{tohyama2004ciair} & En$\leftrightarrow$Jp & 182 \\
EPPS \shortcite{paulik2009automatic} & En$\leftrightarrow$Es & 217 \\
Simul-Trans \shortcite{shimizu2014collection} & En$\leftrightarrow$Jp & 22 \\
\hline\hline
BSTC (ours) & Zh$\rightarrow$En &68\\\hline\hline
\end{tabular}
}
\end{center}
\caption{Existing speech translation corpora and ours. The duration statistics of all datasets are rounded up to an integer hour. For MuST-C, the ``8 Euro langs'' is short for ``8 European languages''. Europarl-ST contains the speech translation between 9 European languages. }
\label{tbl:inst}
\end{table}

To train an AST model, existing corpora can be classified into two categories:
\begin{itemize}
\item{\textbf{\textit{Speech Translation}} corpora consist pairs of audio segments and their corresponding translations.}
\item{\textbf{\textit{Simultaneous Translation}} corpora are constructed by transcribing lecturers' speeches and the streaming utterance of human interpreters.}
\end{itemize}
The main difference between these two kinds of corpora lies in the way that the translations are generated. The translations in \textit{Speech Translation} corpora are generated based on complete audios or their transcripts, while the translations in \textit{Simultaneous Translation} corpora are transcribed from real-time human interpretation.

\begin{figure*}[tbp]
\centering
\includegraphics[width=\linewidth]{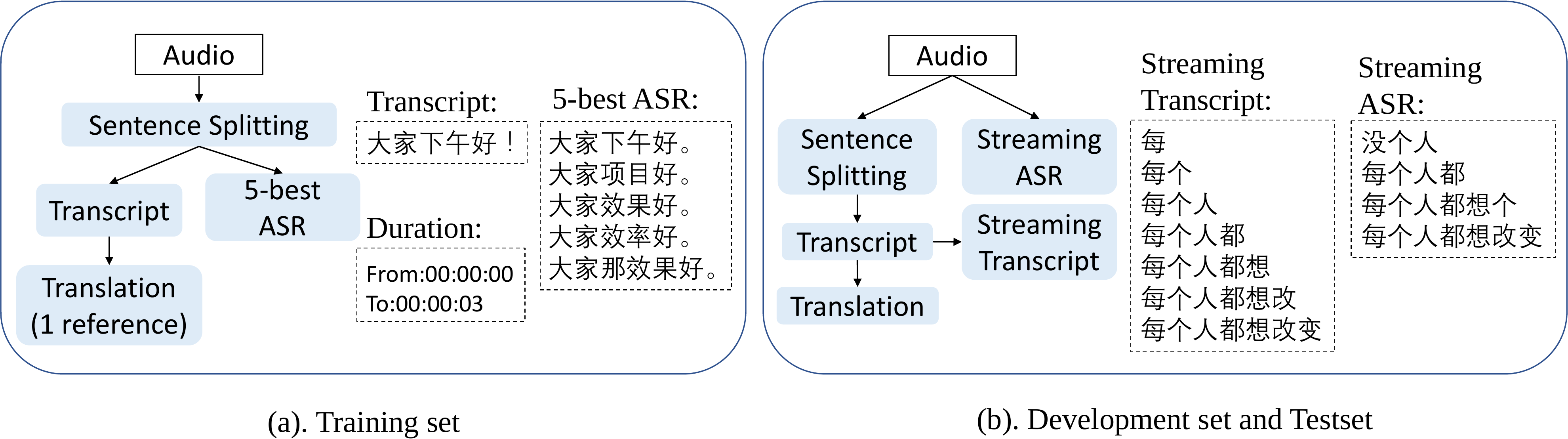}
\caption{The process of constructing the training set and development/test sets (dev/test). The difference between the two processes is that for the training set we first split audio into sentences and then get the ASR and transcript for each sentence, while for the dev/test sets we record the real-time ASR and transcript, the sentence splitting is only used to generate translations of segmented sentences.}
\label{fig:train_dev_test}
\end{figure*}

Existing research on \textit{Speech Translation} mainly focused on the translation between English and Indo-European languages\footnote{Indo-European languages are a large language family.}, with little attention paid to that between Chinese (Zh) and English. One of the reasons is the scarcity of public Zh$\leftrightarrow$En speech translation corpora. Among the public corpora, only MSLT \cite{federmann2017microsoft} and Covost \cite{wang2020covost,wang2020covost2} contains Zh$\leftrightarrow$En speech translation, as shown in Table \ref{tbl:inst}. But the total volume of them on Zh$\rightarrow$En translation is merely about 30 hours, which is too small to train data-hungry neural models. Some studies explore Zh$\rightarrow$En \textit{Simultaneous Translation} \cite{DBLP:journals/corr/abs-1810-08398,zhang-etal-2020-learning-adaptive}. However, they take text translation datasets to simulate real-time translation scenarios because of the lack of simultaneous translation corpus.


To promote the research on Chinese-English speech translation, as well as evaluating the translation quality in real simultaneous interpretation environments, we construct BSTC, a large-scale Zh$\rightarrow$En speech translation and simultaneous translation dataset including approximately 68 hours of Mandarin speech data with their automatic recognition results, manual transcripts, and translations. Our contributions are:
\begin{itemize}
\item We propose the first large-scale (68 hours) Chinese-English \textit{Speech Translation} corpus. This training set is a four-way parallel dataset of Mandarin audio, transcripts, ASR lattices, and translations.
\item The proposed dev and test set constitutes the first high-quality \textit{Simultaneous Translation} dataset of over 3-hour Mandarin speech, together with its streaming transcript, streaming ASR results, and high-quality translation.
\item We have organized two simultaneous interpretation tasks\footnote{We organized two shared tasks on the 1st and 2nd Workshop on Automatic Simultaneous Translation.} to promote research in this field and deployed a strong benchmark on this dataset.
\item The proposed dataset can also be taken as 1) a \textit{Chinese Spelling error Correction} (CSC) corpus containing pairs of ASR results and corresponding manual transcripts or 2) a Zh$\rightarrow$En \textit{Document Translation} dataset with context-aware translations.
\end{itemize}

\begin{table*}[]
\begin{center}
\resizebox{\linewidth}{!}{
\begin{tabular}{l|c|c|c|c|c|c}
\hline
Dataset & Talks & Utterances &Transcription (characters) &Translation (tokens)&Audio (hours) &WER(1-best) \\
\hline
Train &215&37,901&1,028,538&606,584&64.57&27.90\%\\
Dev &16& 956&26,059&75,074 & 1.58&15.21\%\\
Test&6& 975& 25,832&70,503&1.46&10.32\%\\
\hline
\end{tabular}}
\end{center}
\caption{The summary of our proposed speech translation data.}
\label{tbl:data}
\end{table*}

All data can be obtained at the site of our shared task: \url{https://aistudio.baidu.com/aistudio/competition/detail/44} after registration.

\section{Dataset Description}

BSTC is created to fill the gap in Zh$\rightarrow$En speech translation, in terms of both size and quality. To achieve these objectives, we start by collecting approximate 68 hours of mandarin speeches from three TED-like content producers: BIT\footnote{\url{https://bit.baidu.com}}, \textit{tndao.com}\footnote{\url{http://www.tndao.com/about-tndao}}, and \textit{zaojiu.com}\footnote{\url{https://www.zaojiu.com/}}.
The speeches involve a wide range of domains, including IT, economy, culture, biology, arts, etc. We randomly extract several talks from the dataset and divide them into the development and test set.

\subsection{Training set}
For the training set, we manually tag timestamps to split the audio into sentences, transcribe each sentence and ask professional translators to produce the English translations. The translation is generated based on the understanding of the entire talk and is faithful and coherent as a whole. To facilitate the research on robust speech translation, we also provide the top-5 ASR results for each segmented speech produced by SMLTA\footnote{\url{http://research.baidu.com/Blog/index-view?id=109}}, a streaming multi-layer truncated attention ASR model. Figure \ref{fig:train_dev_test} (a) shows the construction process of the training set, together with an example of a segmented sentence.

\begin{figure}[tbp]
\centering
\includegraphics[width=2.6in]{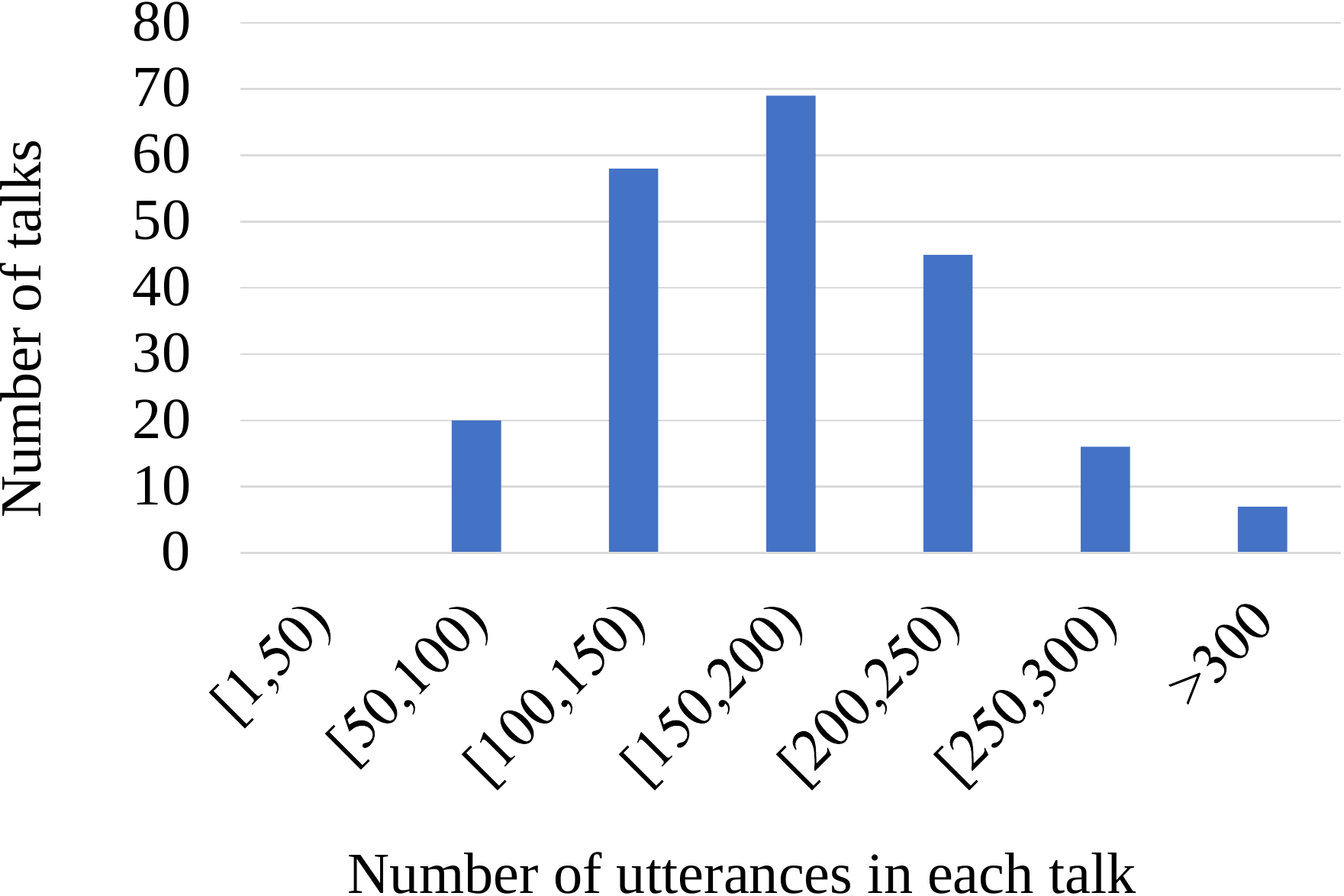}
\caption{The distribution of talk length (number of sentences) in the training set.}
\label{fig:data_nsents}
\end{figure}
\subsection{Dev/Test set}
For the development (dev) set and test set, we consider the simultaneous translation scenario and provide the streaming transcripts and streaming ASR results, as shown in Figure \ref{fig:train_dev_test} (b). The streaming transcripts are produced by turning each $n$-words (a word means a Chinese character here) sentence to $n$ lines word by word with length $1, 2, ..., n$. We use the real-time recognition results of each speech, rather than the recognition of each sentence-segmented audio. This is to simulate the simultaneous interpreting scenario, in which the input is streaming text, rather than segmented sentences.
\begin{figure}[tbp]
\centering
\includegraphics[width=2.8in]{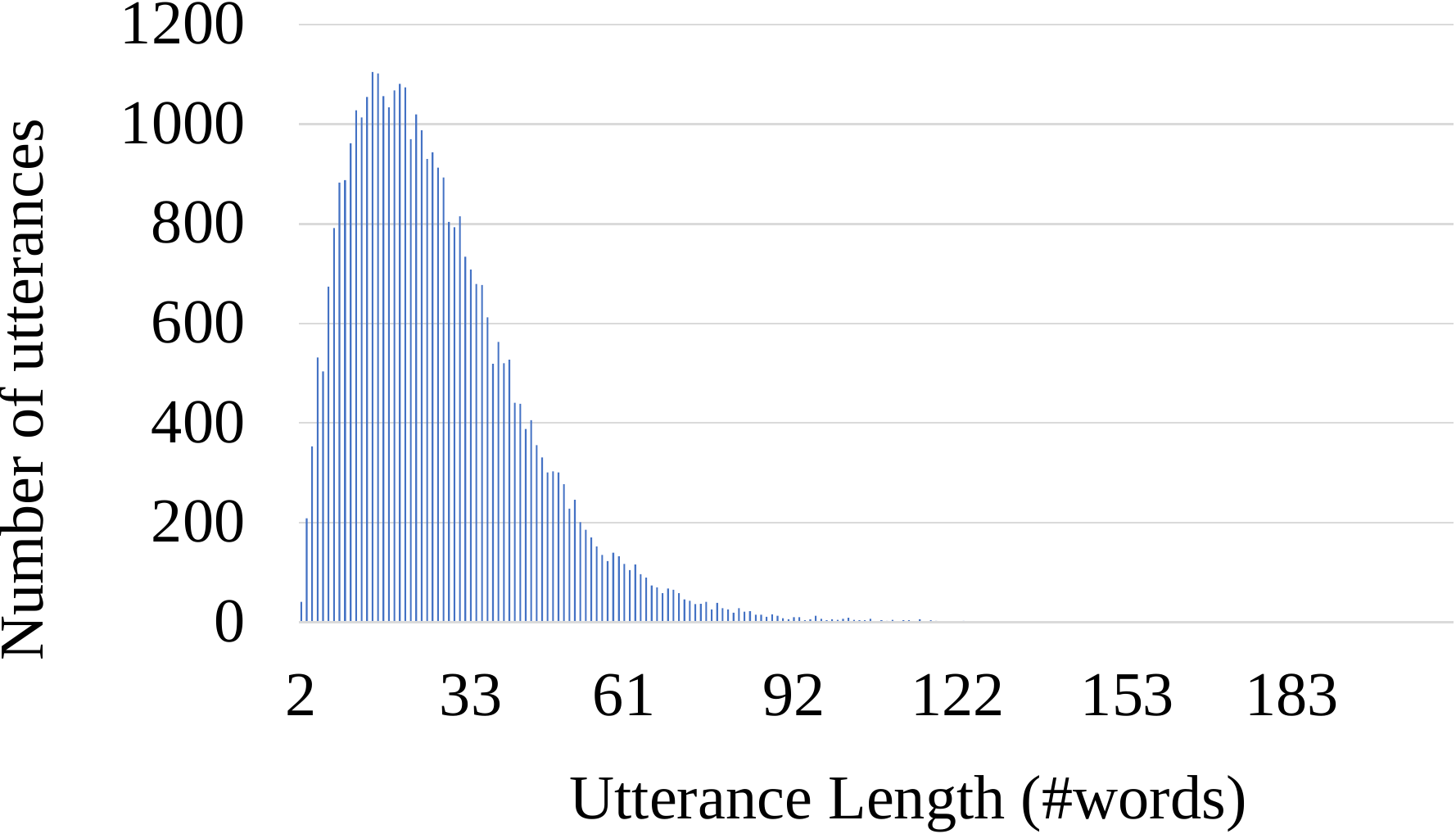}
\caption{The distribution of utterance length (number of words) in the training set. A word means a Chinese character here.}
\label{fig:data_sentlen}
\end{figure}
\subsection{Statistics and Dataset Features}
We summarize the statistics of our dataset in Table \ref{tbl:data}. The distribution of talk length and utterance length in the training set is illustrated in Figure \ref{fig:data_nsents} and Figure \ref{fig:data_sentlen}, respectively. The average number of utterances per talk is 176.3 in the training set, 59.8 in the dev set, and 162.5 in the test set. And the average utterance length is 27.14 in the training set, 27.26 in the dev set, and 26.49 in the test set.

We also calculate the word error rate\footnote{WER tool: \url{https://github.com/belambert/asr-evaluation}} (WER) of the ASR system on the three datasets. As shown in Table \ref{tbl:data}, the WER of the training set is 27.90\%, significantly higher than that of the dev and testset. This is due to the way of audio segmentation before recognition: some audio clips lose some parts in acoustic truncation, resulting in incomplete ASR results. We count the length difference of each $<$transcription, asr$>$ pair, i.e., $\Delta_{len}=|len(transcription)-len(asr)|$, and recalculate the WER of pairs whose length difference is within a certain range. The WER and coverage of these subsets are listed in Table \ref{tbl:train_wer}. Note that when the asr and transcript with equal length ($\Delta_{len}\le0$), the WER is only 5.87\%. For the length difference in a relatively regular range (e.g, $\Delta_{len}\le15$), the WER is also relatively low (WER=15.23\%).

\begin{table}
\begin{center}
\resizebox{4.3cm}{!}{
\begin{tabular}{l|c|c}
\hline
$d_{len}$ & WER & Coverage \\ \hline
0 & 5.87\% & 31.61\% \\
1 & 7.13\% & 55.30\% \\
3 & 8.86\% & 68.50\% \\
7 & 10.72\% & 74.50\% \\
15 & 15.23\% & 83.40\% \\
31 & 23.51\% & 94.00\% \\
∞ & 27.90\% & 100\% \\ \hline
\end{tabular}}
\end{center}
\caption{The WER and coverage of different subsets of the training set with the length difference $\Delta_{len}$ between transcript and asr lower than or equal to $d_{len}$.}
\label{tbl:train_wer}
\end{table}

\begin{table}
\centering
\resizebox{5.3cm}{!}{
\begin{tabular}{l|c|c|c}
\hline
& \multicolumn{1}{l|}{BLEU} & AP & Omissions \\
\hline
A & 24.20 & 83.0\% & 53\% \\
B & 17.14 & 62.8\% & 47\% \\
C & 25.18 & 76.5\% & 53\% \\
\hline
\end{tabular}}
\caption{Comparison of the simultaneous interpretation results of three interpreters (A, B, and C) on the BSTC test set. ``AP'' is the Acceptability and the ``Omissions'' indicates the proportion of missing translation in all translation errors.}
\label{tbl:human}
\end{table}

%

Besides, there is a difference between our dataset and the existing speech translation corpora. In our dataset, speech irregularities are kept in transcription while omitted in translation (eg. filler words like ``嗯, 呃, 啊'', unconscious repetitions like ``这个这个呢'' and some disfluencies), which can be used to evaluate the robustness of the NMT model dealing with spoken language. Some other large-scale speech translation datasets \cite{kocabiyikoglu2018augmenting,di2019must}, on the contrary, ignore these speech irregularities in the transcript.

\subsection{Human Interpretation}
We further ask three experienced interpreters (A, B, and C) with interpreting experience ranging from four to nine years to interpret the six talks of the testset, in a mock conference setting\footnote{We play the video of the speech, just like in a real simultaneous interpretation scene}.

To evaluate their translation quality, we also ask human translators to evaluate the transcribed interpretation from multiple aspects: adequacy, fluency, and correctness:
\begin{itemize}
\item \textbf{Rank1}: The translation contains no obvious errors.
\item \textbf{Rank2}: The translation is comprehensible and adequate, but with minor errors such as incorrect function words and less fluent phrases.
\item \textbf{Rank3}: The translation is incorrect and unacceptable.
\end{itemize}

\begin{figure}[tbp]
\centering
\includegraphics[width=\linewidth]{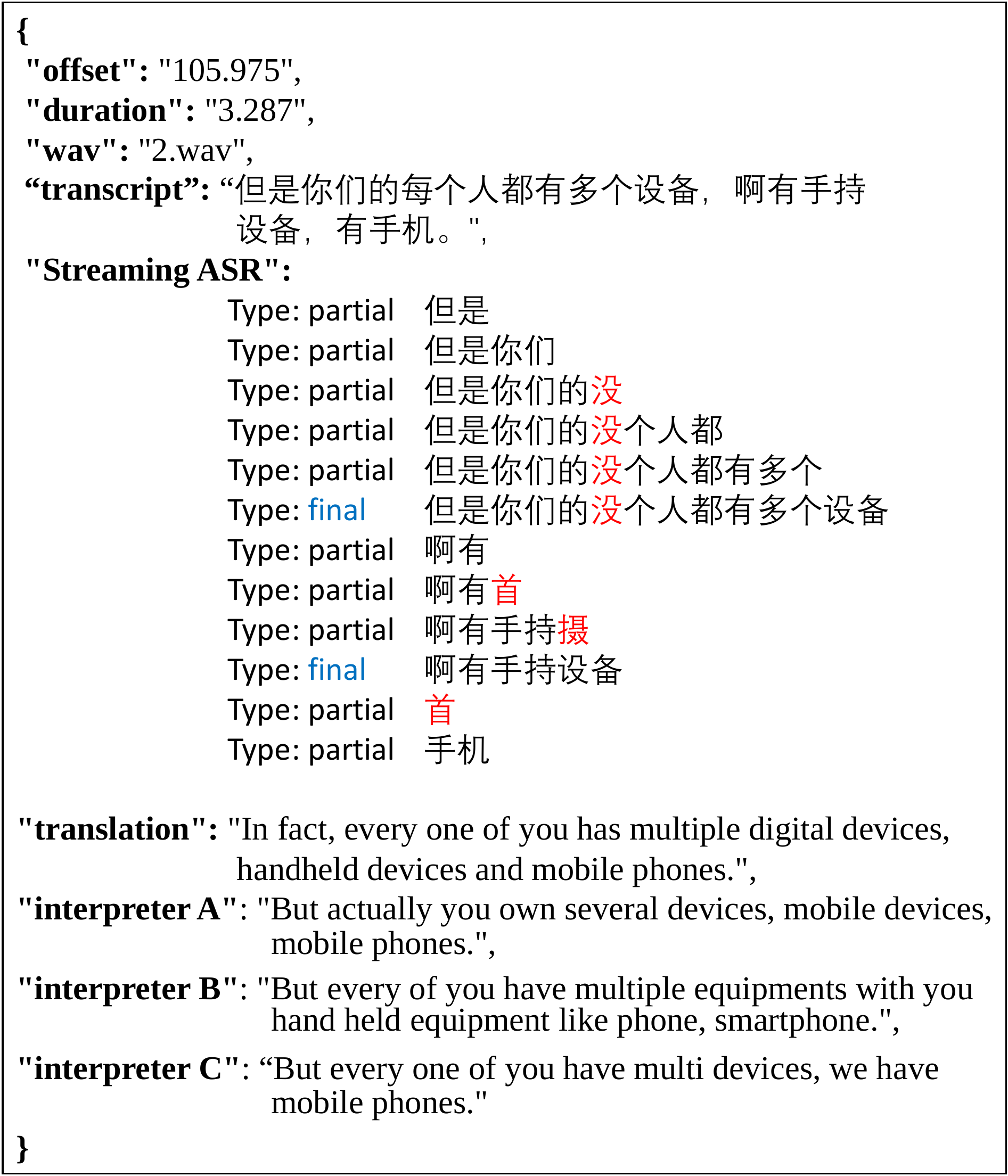}
\caption{A segment of one example in our test set，including audio, timelines, transcription, translation, streaming ASR results, and interpretation from three human interpreters (only for testing data). The red characters in ``Streaming ASR'' indicate recognition errors.}
\label{fig:sample}
\end{figure}
\begin{table*}
\centering
\resizebox{15cm}{!}{
\begin{tabular}{l|c|c|c|c}
\hline
\multirow{2}{*}{Systems} & \multicolumn{2}{c|}{\begin{tabular}[c]{@{}c@{}}Test on Transcript\end{tabular}} & \multicolumn{2}{c}{\begin{tabular}[c]{@{}c@{}}Test on ASR\end{tabular}} \\
\cline{2-5}
& Dev & Test & Dev & Test \\
\hline
pre-train on WMT & 20.78 & 35.13 & 18.22 & 33.32 \\
~ ~ ~ ~Finetune on $<$transcript, translation$>$ & \textbf{23.47}(2.69$\uparrow$) & \textbf{41.14}(6.01$\uparrow$) & 19.68(1.46$\uparrow$) & 35.71(2.39$\uparrow$) \\
~ ~ ~ ~Finetune on $<$ASR, translation$>$ & 22.53(1.75$\uparrow$) & 39.23(4.1$\uparrow$) & \textbf{19.82}(1.6$\uparrow$) & \textbf{36.89}(3.57$\uparrow$) \\
\hline
\end{tabular}}
\caption{The results of benchmark trained on different training datasets, and evaluated by streaming transcription and ASR input.}
\label{tbl:overall}
\end{table*}

Table \ref{tbl:human} shows the translation quality in BLEU and acceptability, which is calculated as the sum of the percentages of Rank1 and Rank2. It shows that their acceptability ranges from 62.8\% to 83.0\%, but the acceptability and BLEU are not completely positively correlated. This is because human interpreters routinely omit less important information to overcome their limitations in working memory. Acceptability focuses more on accuracy and faithfulness than adequacy, so it can tolerate information omission. Therefore, some information omitted in human interpretation that results in inferior BLEU may not lead to the decrease of acceptability. But BLEU, as a statistical auto-evaluation metric, considers adequacy with the same importance with accuracy. This leads to the discrepancy between BLEU and acceptability.

Figure \ref{fig:sample} lists a segment from one example in our dataset. Notably, we only supply human interpretations for testing data. Here the ``Streaming ASR'' is the real-time recognition results, in which the ``Type:final'' means that the audio has detected a pause or silence and thus segmented, and will start to recognize a new sentence, while ``Type:partial'' is to continue recognizing the current sentence.

\section{Experiments}
In this section, we introduce our benchmark systems based on the dataset. We conduct experiments on speech translation and simultaneous translation, respectively.

To preprocess the Chinese and the English text, we use an open-source Chinese Segmenter\footnote{\url{https://github.com/fxsjy/jieba}}, and Moses Tokenizer\footnote{\url{https://github.com/moses-smt/mosesdecoder/blob/master/scripts/tokenizer/tokenizer.perl}}. After tokenization, we convert all English letters into lower case.
To train the MT model, we conduct byte-pair encoding \cite{DBLP:journals/corr/SennrichHB15} for both Chinese and English by setting the vocabulary size to 20K and 18K for Chinese and English, respectively. And we use the ``multi-bleu.pl'' \footnote{\url{https://github.com/moses-smt/\\mosesdecoder/blob/master/scripts/generic/multi-bleu.perl}} script to evaluate the BLEU score.

\subsection{Benchmark System}
Our benchmark is a cascade system that includes an ASR module, a sentence segmentation module, and a machine translation (MT) module.

\begin{itemize}
\item We use the SMLTA model for ASR, i.e., the streaming transcript/ASR of BSTC is taken as the output of the ASR module.
\item The sentence segmentation module is to decide when to translate in real-time. We train a classification model based on the Meaningful Unit (MU) method proposed in \citet{zhang-etal-2020-learning-adaptive} that implements a 5-class classification (MU, comma, period, question mark, and none). The training data of meaningful units are generated automatically from monolingual sentences based on context-aware translation consistency. The model is pre-trained on ERNIE-base \cite{sun2020ernie} and fine-tuned on the transcript of the BSTC training set.
\item Once an MU or a sentence boundary (period or question mark) is detected in the sentence segmentation module, the MT module generates translation for the detected sentence. The MT model is firstly pre-trained on the large-scale WMT19 Chinese-English corpus, then fine-tuned on BSTC. The WMT19 corpus includes 9.1 million sentence pairs collected from different sources, \textit{i.e.}, Newswire, United Nations Parallel Corpus, Websites, etc. We use the \textit{big} version of Transformer model in the following experiments.
\end{itemize}





\subsection{Performance of Speech Translation}
Speech translation aims at translating accurately without considering system delay. Therefore, we only perform translation when sentence boundaries (periods and question marks) are detected by the sentence segmentation module. 

The MT model is firstly trained on WMT, then fine-tuned on 37,901 training pairs of $<$transcription, translation$>$ and $<$asr, translation$>$ in two settings, respectively. The purpose of fine-tuning on transcription is to adapt the model to the speech domain, and the purpose of fine-tuning on ASR is to improve the robustness of the MT model against recognition errors. Our model pre-trained on WMT19 achieves a BLEU of 25.1 on Newstest19.

We evaluate our systems on the dev/test set using streaming transcription and streaming ASR as inputs. For each talk in the dev/test set, its streaming text is firstly segmented by the sentence segmentation module, then the translation of each segmentation is concatenated into one long sentence to evaluate the BLEU score. The results are listed in Table \ref{tbl:overall}. Note that the great gap of BLEU in dev and test sets is that, the dev set has only one reference while the testset has 4 references.

\begin{figure*}[t]
\centering
\includegraphics[width=13.5cm]{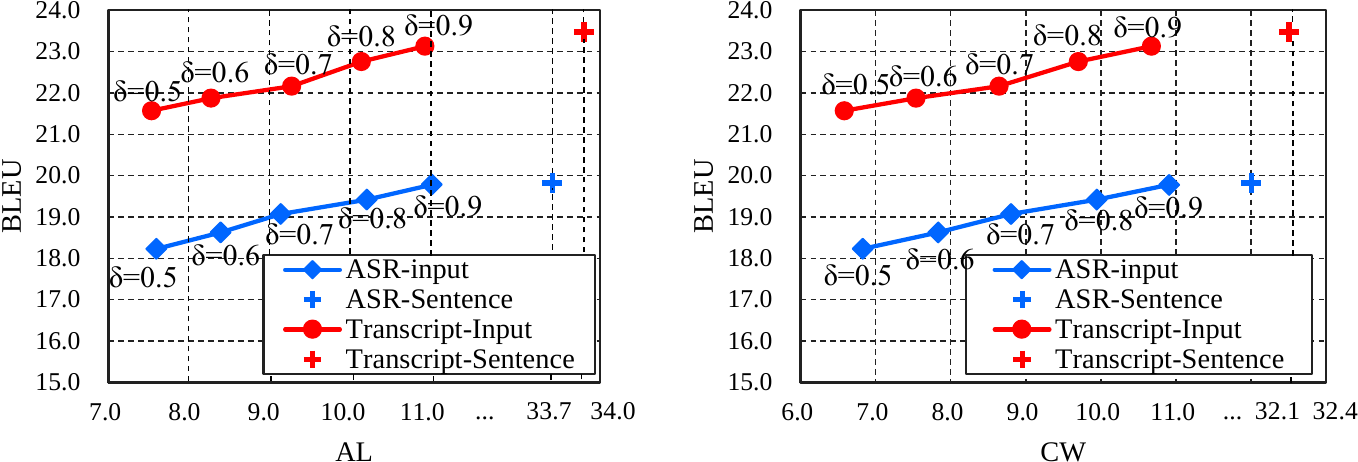}
\caption{Translation quality against latency metrics on BSTC development set. ``ASR-Sentence'' and ``Transcript-Sentence'' denotes the results of full-sentence translation with ASR input and transcript input, respectively. }
\label{fig:simul_dev}
\end{figure*}
\begin{figure*}
\centering
\includegraphics[width=13.5cm]{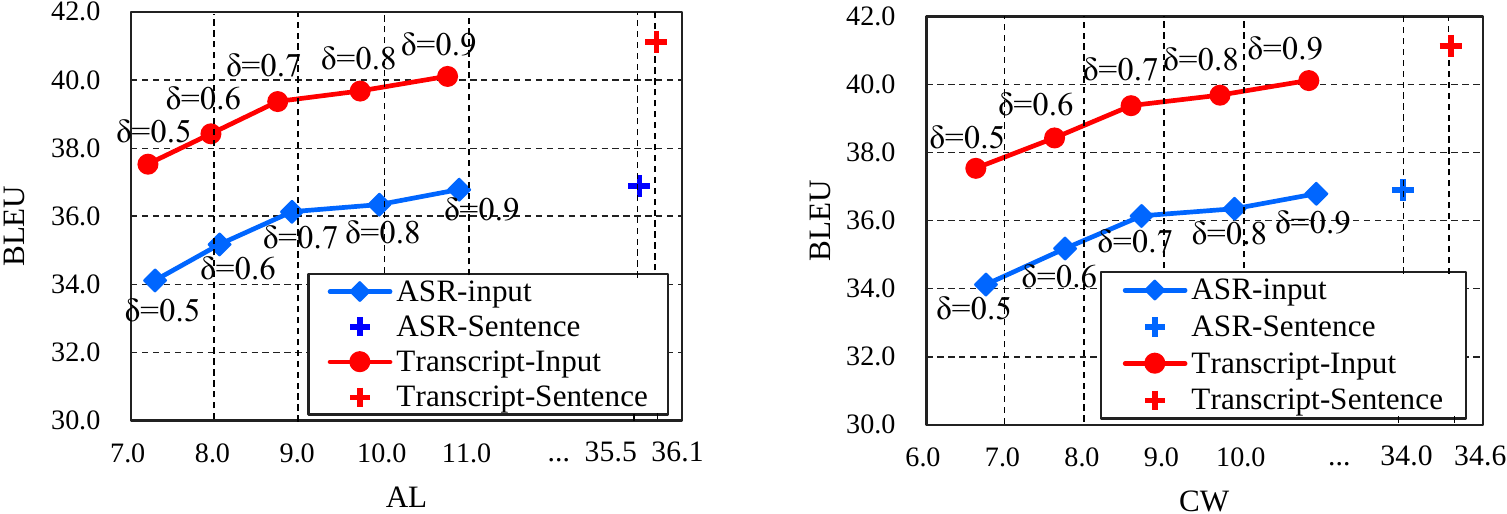}
\caption{Translation quality against latency metrics on BSTC testset.}
\label{fig:simul_test}
\end{figure*}

\noindent \textbf{Contribution of fine-tuning on speech translation data:} The systems pre-trained on WMT obtain an absolute improvement both on clean and noisy input by fine-tuning on $<$transcription, translation$>$. The performance of the former model increases by 4.35 BLEU score on average and the latter model obtains 1.93 BLEU score improvement on average. This indicates the transcribed training data can still bring large improvement after pre-training on large-scale training corpus. This probably because it is closer to the test set in terms of the domain (speech) and noise (disfluencies in spoken language).

\noindent \textbf{Contribution of fine-tuning on noisy data}: Training on the corpus containing the ASR errors can be effective to improve the robustness of the NMT model. This can be proved by fine-tuning on the $<$ASR, translation$>$ pairs. As shown in the last row of Table \ref{tbl:overall}, the pre-trained model improves 2.93 and 2.59 BLEU scores on average for testing on streaming transcript and streaming ASR, respectively. This manifests that compared with fine-tuning the clean transcription, the model fine-tuned on ASR is less sensitive to false recognition results of ASR.

\subsection{Performance of Simultaneous Translation}

Different from speech translation, the simultaneous translation should balance translation quality and latency. Therefore, we fix the ASR and MT modules to evaluate our system under different sentence segmentation results. In simultaneous translation, once an MU or a sentence boundary is detected, the MU or sentence is translated immediately. In order to maintain coherent and consistent paragraph translation, we perform context-aware translation following \citet{xiong2019dutongchuan} that except for the first segment in a sentence, the subsequent segments are translated with force-decoding.

The performance of system on the dev set and test set is listed in Figure \ref{fig:simul_dev} and Figure \ref{fig:simul_test}, respectively\footnote{We list detailed values in Table \ref{tbl:specific}}. We use BLEU to evaluate the translation quality and use average lagging (AL) \cite{DBLP:journals/corr/abs-1810-08398} and Consecutive Wait (CW) \cite{gu2017learning} as latency metrics. $\delta$ is the hyperparameter defined in \citet{zhang-etal-2020-learning-adaptive} as the thresold of sentence segmentation module. 
It shows that the translation quality improves consistently with the increase of latency. The AL on both dev and test sets ranges from 7 to 12 and the CW ranges from 6 to 11 for points of simultaneous translation.  In addition,  we also draw the full-sentence translation results, as denoted by ``ASR-Sentence'' and ``Transcript-Sentences'' in the two figures. The full-sentence translation implements a high-latency policy, in which a translation is only triggered when a sentence is received. As shown in the figures, the delay of both ``ASR-Sentence'' and ``Transcript-Sentences'' is much higher than the simultaneous translation results.

\begin{table}[t]
\centering
\resizebox{6cm}{!}{
\begin{tabular}{|l|l|l|l|l|}
\hline
                                                & $\mathcal{\delta}$    & \textbf{AL}      & \textbf{CW}      & \textbf{BLEU}   \\ \hline
\multirow{12}{*}{\textbf{Dev Set}}                       & \multicolumn{4}{l|}{\textbf{Input ASR}}        \\ \cline{2-5} 
                                                & 0.5      & 7.61    & 6.82    & 19.07  \\ \cline{2-5} 
                                                & 0.6      & 8.42    & 7.83    & 19.42  \\ \cline{2-5} 
                                                & 0.7      & 9.17    & 8.80    & 19.78  \\ \cline{2-5} 
                                                & 0.8      & 10.26   & 9.94    & 20.25  \\ \cline{2-5} 
                                                & 0.9      & 11.08   & 10.91   & 20.37  \\ \cline{2-5} 
                                                & \multicolumn{4}{l|}{\textbf{Input Transcript}} \\ \cline{2-5} 
                                                & 0.5      & 7.54    & 6.58    & 21.87  \\ \cline{2-5} 
                                                & 0.6      & 8.30    & 7.54    & 22.16  \\ \cline{2-5} 
                                                & 0.7      & 9.31    & 8.64    & 22.76  \\ \cline{2-5} 
                                                & 0.8      & 10.19   & 9.70    & 23.13  \\ \cline{2-5} 
                                                & 0.9      & 11.00   & 10.67   & 23.62  \\ \hline
\multicolumn{1}{|c|}{\multirow{12}{*}{\textbf{Test set}}} & \multicolumn{4}{l|}{\textbf{Input ASR}}        \\ \cline{2-5} 
\multicolumn{1}{|c|}{}                          & 0.5      & 7.28    & 6.75    & 34.12  \\ \cline{2-5} 
\multicolumn{1}{|c|}{}                          & 0.6      & 8.04    & 7.75    & 35.18  \\ \cline{2-5} 
\multicolumn{1}{|c|}{}                          & 0.7      & 8.90    & 8.71    & 36.14  \\ \cline{2-5} 
\multicolumn{1}{|c|}{}                          & 0.8      & 9.93    & 9.88    & 36.35  \\ \cline{2-5} 
\multicolumn{1}{|c|}{}                          & 0.9      & 10.87   & 10.91   & 36.79  \\ \cline{2-5} 
\multicolumn{1}{|c|}{}                          & \multicolumn{4}{l|}{\textbf{Input Transcript}} \\ \cline{2-5} 
\multicolumn{1}{|c|}{}                          & 0.5      & 7.20    & 6.62    & 37.54  \\ \cline{2-5} 
\multicolumn{1}{|c|}{}                          & 0.6      & 7.94    & 7.61    & 38.43  \\ \cline{2-5} 
\multicolumn{1}{|c|}{}                          & 0.7      & 8.73    & 8.58    & 39.38  \\ \cline{2-5} 
\multicolumn{1}{|c|}{}                          & 0.8      & 9.70    & 9.70    & 39.69  \\ \cline{2-5} 
\multicolumn{1}{|c|}{}                          & 0.9      & 10.74   & 10.81   & 40.12  \\ \hline

\end{tabular}
}
\caption{Specific data corresponding to Figure \ref{fig:simul_dev} and Figure \ref{fig:simul_test}.}
\label{tbl:specific}
\end{table}

\section{Conclusion and Future Work}
In this paper, we release a challenging dataset for the research on Chinese-English speech translation and simultaneous translation. Based on this dataset, we report a competitive benchmark based on a cascade system. In the future, we will expand this dataset, and propose an effective method to develop an End-to-End speech translation model.

\bibliography{dataset}
\bibliographystyle{acl_natbib}

\end{CJK*}
\end{document}